\documentclass{article}

\author{ Mohammadamir Kavousi       \and Sepehr Saadatmand \\
Southern Illinois University, Carbondale, USA \\
\{mohammadamir.kavousi, sepehr.saadatmand\}@siu.edu
}

\date{}

\input{Definitions.sty}
\title{Estimating the Rating of Reviewers Based on the Text}
\begin{document}
\maketitle

\section{Abstract}
User-generated texts such as reviews and social media are valuable sources of information. Online reviews are important assets for users to buy a product, see a movie, or make a decision. Therefore, rating of a review is one of the reliable factors for all users to read and trust the reviews. This paper analyzes the texts of the reviews to evaluate and predict the ratings. Moreover, we study the effect of lexical features generated from text as well as sentimental words on the accuracy of rating prediction. Our analysis show that words with high information gain score are more efficient compared to words with high TF-IDF value. In addition, we explore the best number of features for predicting the ratings of the reviews. \\
\textbf{Keywords:} Review Mining, Natural Language Processing, Machine Learning, Big Data.


\section{Introduction}
With a rapid growth of Internet and online shopping systems, customers share opinion on online platforms to assist other customers in making wiser decisions. This contribution has resulted in active communities which are known as valuable sources for both scholars and industry owners. Online reviews are important assets for users to buy a product, see a movie, or choose a product. Moreover, ratings of the reviews are important factors for showing the quality of a product, and as the result, it can be used as a reliable feature for the users to read and trust the reviews or purchase a product \cite{zhang2015product}. Review ratings are also used in recommender systems to push and automatically suggest products to users, based on similar choices and attributes compared to others. Due to their importance, business owners and academic scholars have studied user generated reviews to find efficient techniques for estimating ratings based on the content of the text \cite{bagherzadeh2016investigating, wilson2005recognizing, kim2006automatically, de2011joint, danescu2009opinions}. The research in the area of review mining is very vast and is not just limited to finding rating analysis. Other areas such as opinion extraction and sentiment analysis, building recommendation systems and summarizing the texts are among domains that are extensively explored in this area \cite{pang2005seeing, cui2006comparative, liu2012survey, liu2012movie, hu2004mining, jindal2008opinion,li2010structure} .

Various numbers of studies focused on predicting ratings of different products on Amazon (please refer to section 2 for more information). In this paper, we focus on predicting the ratings of films (documentary and non-documentary) based on the text of the reviews. Documentary films have reviews with richer texts focusing on the themes of the films, while non-documentary films consist of attributes such as famous casts. We believe that a combination of these two types of films and using the textual features will provide good insights about the underlying characteristics of the texts.

The rest of the paper is organized as follows: section 2 discusses related work. Section 3 discusses the data collection.  In the method section (4), we first explain the feature selection and discuss the classifiers that we chose for this study, and then report the results of the classifier. Section 5 discusses the conclusion and future directions for improving the paper.

\section{Related Works}
As mentioned earlier prior work on review mining is very vast. Researchers in this area have tried to find efficient algorithms for predicting rating, helpfulness, and sentiment of the reviews \cite{haddadpour2016low, fahim2017optimal, hong2012reviews, wang2010latent, ravandi2016black, liu2012movie, lu2010exploiting, farhadii}. In this section, we explore the most related work in the area of review mining and analyze the papers that used text of the reviews to extract information. 
Bing and Zhang (2012) reviewed the most efficient algorithms in opinion mining and sentiment analysis of the reviews. Supervised and unsupervised approaches are employed to extract the sentiment of the reviews on sentence or document level \cite{wilson2005recognizing}. Since a positive or negative opinion about a product (e.g. Camera) does not show the feeling of the opinion holder about every specific feature of that product (e.g. picture quality of a camera), aspect-based sentiment analysis is introduced to find the opinions related to each attribute on the sentence level.

In a research, De Albornoz et al. \cite{de2011joint} used both topic and sentiment of the reviews on sentence level to assess the impact of text-driven information in predicting the rating of the reviews in recommendation systems. This article aims to predict the overall rating of a product review based on the user’s opinion about different product features that are evaluated in the review. After identifying the features that are relevant to consumers, they extracted users' opinions about different product features. The salience of different product features and the values that quantify the users' opinions are used to construct the feature vector to represent the review. This vector was used as the input of the machine learning model to classify the review in different rating categories. As a result, the models achieved 84 percent for Logistic, 83 percent for LibSVM and 81.9 percent for FT. The errors are assumed to be the result of: (1) mislabeled instances in the training set (2) frequent spelling errors in the reviews and (3) the presence of neutral sentences which do not express any opinion but are necessarily classiﬁed as positives or negatives. 
In another research, Ghose et al \cite{ghose2007designing} conducted a study to create a dataset of products from the Amazon website. The dataset consisted of product-specific characteristics and the details of the product review. They generated a training set with two classes of documents; a) a set of “objective" documents that contained the product descriptions of each 1,000 products and b) a set of “subjective" documents that contained randomly retrieved reviews. As reported, this model successfully identified the most helpful reviews to the users. The most “helpful” reviews can be displayed on top to improve users’ re-viewing experience on electronic marketplaces \cite{chehardeh2016remote, ghose2011estimating, mozaffari2018generalized, mozaffari2017new,mudambi2010research}.

Kim et al. in \cite{kim2006automatically} suggested an automated assessment for the review helpfulness by considering the length of the review as the most useful feature compared to the other ones. The main contribution of this paper is: a) using helpfulness score to implement an automatic computational model to rank the reviews, and b) leveraging various features in the reviews such as structural, lexical, syntactic, semantic, and reviews’ starts to predict the helpfulness score of the reviews. For products such as MP3 players and cameras, the model achieved correlation coefficient scores of 0.656 and 0.604. The detailed analysis of features showed that length of the reviews, unigrams, and product rating were the most helpful ones in the prediction and structural and syntactic features had no significant impact \cite{zhuang2006movie}.

Rezapour and Diesner in \cite{rezapour2017classification} studied the movie reviews from a new perspective that, as claimed by the authors, is new in the area of review and opinion mining. In this work, the authors captured the impact of movies from the reviews by first creating a novel dictionary of impact and then annotating the reviews on sentence level with various types of impact as change in behavior, change in cognition, and etc. They used three different classifiers with three sets of features to classify the impact of each film on the authors. The results showed that SVM classifier and the combination of all features are the best predictor of impact in reviews. In another work, the amount of alignment of social media (Facebook and reviews), news articles and the transcript of the film were studied \cite{diesner2016assessing}. It was found that social media are more aligned with the main subject of the films. The result of these two studies show that films are important sources that are capable of influencing people’s behavior and cognition.

The study presented in this paper builds upon the previous research. We study the influence of lexical features as well as the impact of feature size on the classification task. Moreover, our analysis will highlight the importance of using the correct size for rating classification. Note that this study is a small-scale rating prediction. Using the insights of this paper, for data parallel processing of big data, more sophisticated algorithms based on MapReduce can be used for speeding up the processing time, e.g. look at \cite{wang2016maptask, yekkehkhany2017near, daghighi2017scheduling, yekkehkhany2018gb, kavousi2017affinity}.

\section{Data}
We used the reviews of eight films from Amazon . Table \ref{table:1} shows the names and the number of the reviews for each film.  Around 65 percent of the reviews are 5 star and 20 percent of them are 1, 2 and 3 stars. To have more distinct classes, 4-star reviews were excluded from the dataset. For preprocessing, first, all the reviews were divided in two categories/classes as High (5Stars) and Low (1, 2, and 3 stars). We then removed the stop words and tokenized the sentences. All the preprocessing was performed using Python NLTK \cite{bird2009natural} and custom programs. The resulted dataset consisted of around 39,802 sentences and 307,138 words. Words in a collection of documents mostly follow the zip’s law, where the rare and common words are scattered on the end of two tails of the graph. To address this problem, we removed the words with less than 10 counts. The words with high counts will be downscaled in feature selection, using TF-IDF or information gain. More details can be found in the following section. 

\begin{table}
\caption{Number of reviews of each movie}
\label{table:1}

\begin{center}
	\begin{tabular}{ |c|c|c|c|c|c|c| } 
		\hline
		Name &  Reviews &5 Star&4 Star &3 Star&	2 Star &1 Star \\ 
		\hline
		Food Inc.&	2462&	1949&	357&	78&	30&	48 \\
		\hline
		Boyhood	& 2253&	872&	342	&318&	296&	425\\
		\hline
		Fed Up&	1401&	1069&	185&	82&	29&	36\\
		\hline
		Blackfish&	955	&750&	117&	42&	15&	31\\
		\hline
		The Imitation Game&	829&	577	&158&	54&	14&	26\\
		\hline
		Super-Size Me&	670&	324&	152&	71	&41	&82\\
		\hline
		Inside Job&	437&308&	54&	21&	10&	44\\
		\hline
		Citizenfour &	199&	168&	17	&10	&1&	3 \\
		 
		\hline
	\end{tabular} 
\end{center}
\end{table}

\section{Methodology}
Reviews as users generated texts entail feelings and personal ideas of the users \cite{rezapour2017classification}. Unlike usual opinion mining and sentiment analysis in the field of review mining, movies such as documentary films do not benefit from special attributes such as famous cast or directors. Since the aim of a documentary movie is to raise the awareness about a social issue or introduce a new topic, individuals try to focus on these areas in their reviews as well \cite{rezapour2017classification}. To find the ratings of the reviews, some prior studies leveraged helpfulness level as one of the features in rating classification. The helpfulness level is also a great deal in creating the recommendation systems. We did not consider this feature in our study for two reasons: 1) documentary films are not among popular genres of films, and therefore the number of reviews written and viewed by customers is limited. In some cases, this lack of interest results in limited number of helpfulness rate. Moreover, the number of reviews which were viewed more than one time and also had helpfulness rate were around 1400, which is a very small input data for the classifiers. 2) In this study, we just focus on textual features with no external input. 

\subsection{Feature Selection}
\paragraph{TF-IDF.} Term frequency-inverse document frequency (TF-IDF) is a numerical score that can highlight the importance of a word in a document collection or corpus. Areas such as information retrieval and text mining use this score as a weighting factor. TF-IDF score is proportional to the number of times that a word appears in the document. Using this feature, we can downscale the words with high frequency, also known as stop words.  We considered top 500 and 900 words with the highest TF-IDF score to train and test the classifiers.  
We considered top 500 and 900 words with the highest TF-IDF to train the classifiers. 
\paragraph{Information Gain.} This metric leverages the presence or absence of the terms in the documents to calculate the amount of information obtained for each category prediction. To leverage this feature, we calculated the information gain of all words and chose top 200, 600, 900 and 1000 words as features. We compare the results of the classifiers using different sets of features to find the ones that help the most in predicting the ratings.

\paragraph{Sentiment.} One of the popular features in estimating the reviews is using sentiment of the words. In sentiment analysis, each word is tagged with a polarity as positive, negative or neutral. There are two well-known approaches for analyzing the sentiment of the texts. In this paper, we considered a lexicon based approach which leverages a predefined lexicon consisting words as well as their polarity (as a tag or ratio). To get the sentimental words, we used MPQA subjectivity lexicon \cite{wilson2005recognizing}, and extracted and tagged the words in the reviews. In total 2,055 words were extracted. 

After extracting the words of each feature set, we created the feature vectors using python scikit-learn library \cite{pedregosa2011scikit}. Table \ref{table:2} shows the list of features that were used in this study. In addition, top 10 words of each feature set, as Info-Gain, TF-IDF and sentiment are listed in Table \ref{table:3}.
\begin{table}
\caption{List of chosen features}
\label{table:2}
\begin{center}
	\begin{tabular}{ |c|c| } 
		\hline
	 	TF-IDF&	Top 500 words \\ 
	 	TF-IDF&	Top 900 words\\
	 	Info-Gain&	Top 200 words\\
	 	Info-Gain&	Top 600 words\\
	 	Info-Gain	&Top 900 words\\
	 	Info-Gain&	Top 1000 words\\
	 	Sentiment word	&Top sentiment words in more with a count of 5 or more\\	
		\hline
	\end{tabular}
\end{center}
\end{table}
\begin{table}
\caption{List of the top words in selected features}
\label{table:3}
\begin{center}
	\begin{tabular}{ |c|c|c| } 
		\hline
		TF-IDF	&Info-Gain&	Sentiment\\
		\hline
		Unethical&	SeaWorld&	Heavily\\
		Origin&	Boyhood	&Praise\\
		Slim	&Concept&	Amazing\\
		Oversight&	Interesting&	Hard\\
		Marijuana&	Great	&Cruel\\
		Burned&	Documentary	&Attraction\\
		Juices&	Arquette&	Innocent\\
		Sharks	&McDonalds	&Wired\\
		ingest&	Sea	&Hard\\
		investigate&	Opening&	Pure\\
		\hline
	\end{tabular}
\end{center}
\end{table}
\subsection{Classifier}
We used two well-known classifiers, Support Vector Machines (SVM) and Naïve Bayes (NB) to classify the ratings \cite{ganu2009beyond, pang2005seeing}. These two algorithms are among the most common ones in this area of research. Support vector machines are universal learners and are based on the structural risk minimization principle from computational learning theory. In general, SVMs are highly accurate, and will work the best when using an appropriate kernel, especially when the data is not linearly separable. These models also work very well with high-dimensional spaces. Therefore, based on theoretical evidences, SVMs should perform well for text categorization. Naïve Bayes is one of the simple classifiers. They perform well with semi-supervised learning or fully supervised classifications. 
After creating the feature vectors, we randomly selected 90 percent of the data for training the classifiers. The rest of the data will be used for testing the classifiers with the highest accuracy and the most efficient feature sets. We used WEKA \cite{hall2009weka} to implement the two algorithms. Table \ref{table:4} shows the results of the selected features and the values of average accuracy, precision, recall, and f-score of two classifiers, SVM and NB. 
Based on the results in Table \ref{table:4} the highest accuracy was resulted using top 600 Info-Gain words and top 900 Info-Gain words for SVM and top 200 Info-Gain for NB. The average accuracy value, 82.0 percent, and the F-score value, 90 percent are the highest among all other features. SVM classifier achieved a better performance compared to NB. 
\begin{table}
\caption{Results of classifiers }
\label{table:4}
\begin{center}
	\begin{tabular}{ |c|c|c|c|c|c|c|c|c| } 
		\hline
		 \multicolumn{6}{|c|}{ \qquad SVM} & \multicolumn{3}{|c|}{NB} \\
		
		 \hline
		Features&	Acc&	P&	R&	F&	Acc&	P&	R&	F\\
		500  TF-IDF&	50.58&	95&	77&	84&	50.43&	99&	77&	86\\
		900 TF-IDF&	51.17&	94&	77&	84&	50.48&	99&	77	&86\\
		200 Info-Gain&	78.48&	91&	89&	88&	71.86&	86&	87&	85\\
		600 Info-Gain&	81.9&	90&	92&	90&	71.05&	87&	86&	89\\
		900 Info-Gain&	81.61&	90&	91&	90&	70.65&	87&	86&	89\\
		1000 Info-Gain&	77.84&	90&	89&	89&	71.39&	88&	86&	87\\
		Sentiment& 	75.60&	87	&89	&87	&71.04&	84&	87&	85\\
		\hline
	\end{tabular}
\end{center}
\end{table}

We can also see that using information gain is more efficient than sentiment and TF-IDF. The precision value of TF-IDF is very high, but unfortunately, it is just predicting the high-rank reviews which are the larger class. The average accuracy of both top 500 and 900 TF-IDF is the lowest among all others and as Table \ref{table:5} shows the classifiers are not showing enough confidence in predicting the classes. Based on the average accuracy, prediction confidence and F-score values in Table \ref{table:4} and Table \ref{table:5} both top 600 and top 900 information gain words are the best features. Unfortunately, sentiment of the words did not help us in rating as we expected, but compared to TF-IDF are still among the top features.
\begin{table}
\caption{Prediction confidence of the classifiers}
\label{table:5}
\begin{center}
	\begin{tabular}{ |c|c|c| } 
		\hline
		Features&	Confidence& 	Confidence \\
		\hline
		Top 500 TF-IDF&	1.17&	0.8\\
		Top 900 TF-IDF&	2.3	&0.96\\
		Top 200 Info-Gain&	62.1&	44.6\\
		Top 600 Info-Gain&	63.9&	42.11\\
		Top 900 Info-Gain&	63.3&	42\\
		Top 1000 Info-Gain	&55&43\\
		Sentiment &	52&	43\\
		
		\hline
	\end{tabular}
\end{center}
\end{table}

Comparing the result of this paper with other papers, and especially with De Albornoz’s work \cite{de2011joint} (83 percent for SVM), we showed that the result of this research is almost comparable with the other works in this area. One important note here is that reviewed works leveraged various types of features as well sentiment. The result presented in this work is solely based on lexical features. 
To take a step further, we tested the trained classifier on the 10 percent test set data (as explained before). The results are slightly different from what we achieved earlier. Table \ref{table:6} shows the features and the result of the classifiers. Since we have a small data set, we decided to choose (1) top 200 words to avoid overfitting, (2) and top 600 words as one of the best features. Same as the training, SVM resulted in higher accuracy compared to NB. However, the 200 info-Gain words performed better.  

\begin{table}
\caption{Result of the test set}
\label{table:6}
\begin{center}
	\begin{tabular}{ |c|c|c| } 
		\hline
		\multicolumn{2}{|c|}{SVM (Overall Accuracy)} & \multicolumn{1}{|c|}{NB (Overall Accuracy)}  \\
		\hline
		Info-Gain-Top 200&	78&	72\\
		Info-Gain-Top 600&	77&	70\\
		
		\hline
	\end{tabular}
\end{center}
\end{table}
\section{Conclusion}
Based on the results in Table \ref{table:4} and \ref{table:5} increasing the number of attributes in features were not helpful in enhancing the prediction of the ratings. One assumption is that the features that are proportional to the size of the data work better than the high or low number of the attributes. The higher number may result in overfitting of the classifier and the low number may not be able to extract the necessary information from the content to classify the reviews. The best features were top 600 and 900 Information gain words with the highest average accuracy, 82 percent, confidence of the classifier, 64 percent, and f-score, 90 percent. Running the test data showed that sometimes the words of the largest group or domain can dominate the selected features like info gain and may result in biased classifiers. 
In addition, we found that TF-IDF is not always the best metric for extracting the most salient words from the document. We showed that words with the highest information scores perform better. As noted in the methodology and results, we did not combine the features to increase the accuracy. We found that the words in different feature sets highly overlap, which would result in overfitting the classifiers.

We plan to explore other algorithms and approaches in the future. With the popularity of the deep learning algorithms, we can test the same approach using word embedding and LSTM or CNN. In addition, we will consider adding other features such as syntactic features, the length of the reviews and n-gram words to our analysis.
Finally, in our future work, we plan to add social media texts as new features to the rating prediction. Same as reviews, social media such as Twitter, consists of user-generated texts. We believe that this new feature can tremendously help in rating prediction of the reviews, especially in sparse matrix situations. Tweets are great sources of user-specific features such as sentiments, hashtags, location, and texts. We plan to extract the text and hashtags related to films and add them as the sentimental and/or topical word to our features to expand and improve the prediction models. Hashtags were used in previous study in social media analysis for topic modeling \cite{yang2014large}, sentiment analysis \cite{rezapour2017identifying}, and opinion mining \cite{lim2014twitter}. Numbers of research leveraged social media information to predict a movie’s success \cite{lehrer2017box}. However, the research on combining these two user-generated texts is not well explored in the area of rating prediction. 

\bibliographystyle{plain}
\bibliography{reference}
\end{document}